\documentclass{article}

\usepackage[final, nonatbib]{neurips_2019}
\usepackage[square,numbers]{natbib}
\usepackage[utf8]{inputenc} 
\usepackage[T1]{fontenc}    
\usepackage{hyperref}       
\usepackage{url}            
\usepackage{booktabs}       
\usepackage{amsfonts}       
\usepackage{nicefrac}       
\usepackage{microtype}      
\usepackage{amsthm}
\usepackage{amsmath}
\usepackage{xfrac}
\usepackage{wrapfig}
\usepackage{bbm}
\usepackage{subfig}

\usepackage{tikz}
\usepackage{verbatim}
\usetikzlibrary{decorations.pathreplacing}

\newcommand{\RR}{\mathbb{R} }

\newcommand{\Exp}[2]{\mathop{{}\mathbb{E}_{#1}} \left [ #2 \right ] }

\newcommand{\Dcal}{\mathcal{D}}
\newcommand{\Ecal}{\mathcal{E}}

\newcommand{\w}{\mathbf{w}}
\newcommand{\uu}{\mathbf{u}}
\newcommand{\vv}{\mathbf{v}}

\newcommand{\loss}{\ell}
\newcommand*{\vvec}[1]{#1}

\newcommand{\Xcal}{\mathcal{X}}
\newcommand{\Ycal}{\mathcal{Y}}

\newcommand{\defemph}[1]{\textbf{#1}}

\usepackage{xcolor}

\newtheorem{theorem}{Theorem}

\newtheorem{definition}[theorem]{Definition}

\title{A Reparameterization-Invariant Flatness Measure for Deep Neural Networks}

\author{%
  Henning Petzka\\
  Lund University, Sweden\\
  \texttt{henning.petzka@math.lth.se}
   \And
  Linara Adilova \\
  Fraunhofer Center for Machine Learning \\
  Fraunhofer IAIS, Germany \\
    \texttt{linara.adilova@iais.fraunhofer.de}
   \And
    Michael Kamp\\
    Monash University, Australia \\
    \texttt{michael.kamp@monash.edu}
    \And
    Cristian Sminchisescu \\
    Lund University, Sweden\\
    Google Research, Switzerland\\
    \texttt{cristian.sminchisescu@math.lth.se}
}

\begin{document}

\maketitle

\begin{abstract}
The performance of deep neural networks is often attributed to their automated, task-related feature construction. It remains an open question, though, why this leads to solutions with good generalization, even in cases where the number of parameters is larger than the number of samples. Back in the 90s, Hochreiter and Schmidhuber observed that flatness of the loss surface around a local minimum correlates with low generalization error. For several flatness measures, this correlation has been empirically validated. However, it has recently been shown that existing measures of flatness cannot theoretically be related to generalization due to a lack of invariance with respect to reparameterizations.
We propose a natural modification of existing flatness measures that results in invariance to reparameterization. 
\end{abstract}

\section{Introduction}
Neural networks (NNs) have become the state of the art machine learning approach in many applications. An explanation for their superior performance is attributed to their ability to automatically learn suitable features from data. In supervised learning, these features are learned implicitly through minimizing the \defemph{empirical error}
$
\Ecal_{emp}(f,S)= \sfrac{1}{|S|} \sum_{(x,y)\in S} \loss(f(x),y)
$
for a \defemph{training set} $S\subset\Xcal\times\Ycal$ drawn iid according to a \defemph{target distribution} $\Dcal:\Xcal\times\Ycal\rightarrow[0,1]$, and a \defemph{loss function} $\loss:\Ycal\times\Ycal\rightarrow\RR_+$. Here, $f:\Xcal\rightarrow\Ycal$ denotes the function represented by a neural network. 
It is an open question why minimizing the empirical error during deep neural network training leads to good generalization, even though in many cases the number of network parameters is higher than the number of training examples. That is, why deep neural networks have a low \defemph{generalization error}
\begin{equation}
\label{eq:defGenError}
    \Ecal_{gen}= \Exp{(x,y)\sim \Dcal}{\loss(f(x),y)}-\frac{1}{|S|} \sum_{(x,y)\in S} \loss( f(x),y) 
\end{equation}
which is the difference between expected error on the target distribution $\Dcal$ and the empirical error on a finite dataset $S\subset\Xcal\times\Ycal$ sampled from $\Dcal$.

It has been proposed that good generalization correlates with flat minima of the non-convex loss surface~\citep{flatMinima, simplifyingByFlat} and this correlation has been empirically validated~\cite{keskarLarge, sensitivityGeneralization, identifyingGenProperties}. 
%
%
However, as~\citet{dinhSharp} remarked, current flatness measures---which are based only on the Hessian of the loss function---cannot theoretically be related to generalization: For deep neural networks with ReLU activation functions, a linear reparameterization of one layer, $\w_l\rightarrow \lambda \w_l$ for $\lambda>0$, can lead to the same network function by simultaneously multiplying another layer by the inverse of $\lambda$, $\w_k\rightarrow \sfrac 1\lambda \w_k,\ k\neq l$. Representing the same function, the generalization performance remains unchanged. However, this linear reparameterization changes all common measures of the Hessian of the loss. This constitutes an issue in relating flatness of the loss curve to generalization.  We propose a novel flatness measure that becomes invariant under layer-wise reparameterization through a multiplication with $||w||^2$. We empirically show that it also correlates strongly with good generalization performance.  

\section{Measures of Flatness of the Loss Curve}
\label{sec:measuresOfFlatness}
Consider a function $f(x,\w)=\psi(\w, \phi(x))=g(\w \phi(x))$, where $\psi$ is the composition of a twice differentiable function $g:\RR^d\rightarrow \Ycal$ and a matrix product with a matrix $\w\in \RR^{d\times m}$, whereas $\phi:\Xcal\rightarrow \RR^m$ can be considered as a feature extractor. For a loss function $\loss:\Ycal\times \Ycal\rightarrow \RR_+$ we let $H(\w)$ denote the Hessian of the empirical error $\sfrac{1}{|S|} \sum_{(x,y)\in S} \loss(f(x,\w),y)$ on a training set $S$ considered as a function on $\w$ and $\lambda^H_{max}(\vvec \w)$ the largest eigenvalue of $H(\w)$.

\begin{definition}
Let $f(x,\w)=g(\w \phi(x))$ be a model with $g:\RR^m\rightarrow \Ycal$ an arbitrary twice differentiable function on a matrix product of parameters $\w$ and the image of $x$ under a (feature) function $\phi$. Then $\kappa^\phi(\w)$ shall denote a flatness measure of the loss curve, defined by $\kappa^\phi(\w) := ||\vvec \w||^2\cdot \lambda^H_{max}(\vvec \w).$ (Note that small values of $\kappa^\phi(\w)$ indicate flatness and high values indicate sharpness.
)
\end{definition}

\paragraph{Linear regression with squared loss}

In the case of linear regression, $f(x,\w)=\w x\in\RR$ ($\Xcal=\RR^d$, $g=id$ and $\phi=id$), and $\loss$ the squared loss function $\loss(\hat y,y)=(\hat y-y)^2$, we can easily compute second derivatives with respect to $\w\in \RR^d$ to be
$\sfrac{\partial^2 \loss}{\partial \hat y^2}=2$ and the Hessian is independent of the parameters $\w$. In this case, $\kappa^{id}=c\cdot ||\vvec \w||^2$ with a constant $c=2\lambda_{max}(\sum_{x \in S}xx^t)$ and the measure $\kappa^{id}$ reduces to (a constant multiple of) the well-known Tikhonov (ridge) regression penalty.


 

\paragraph{Layers of Neural Networks}

We consider neural network functions \begin{equation}f(x)=\w_L \sigma (\ldots \sigma(\w_2 \sigma(\w_1 x+b_1)+b_2)\ldots )+b_L\end{equation} of a neural network of $L$ layers with nonlinear activation function $\sigma$. We hide a possible non-linearity at the output by integrating it in a loss function $\loss$ chosen for neural network training. By letting $\phi^l(x)=\sigma ( \w_{l-1} \sigma (\ldots \sigma(\w_2 \sigma(\w_1 x+b_1)+b_2)\ldots)+b_{l-1})$ denote the output of the composition of the first $l-1$ layers and $g^l(z)=\w_L \sigma (\ldots \sigma(z+b_{l})\ldots )+b_L$ the composition of the activation function of the $l$-th layer together with the rest of layers, we can write for each layer $l$, $f(x,\w_l)=g^l(\w_l \phi^l(x))$. Then, for each layer of the neural network, we obtain a measure of flatness at parameters $\w$ with $\kappa^l(\w):=||\vvec \w_{l}||^2\cdot \lambda_{max}^{H,l}(\vvec{\w}_{l})$
with $\lambda^{H,l}(\vvec{\w}_{l})$ the largest eigenvalue of the Hessian of the loss with respect to the parameters of the l-th layer. 

\begin{theorem}\label{thm:invariance}
Let $f=f(\w_1,\w_2,\ldots, \w_L)$ denote a neural network function parameterized by weights $\w_l$ of the $l$-th layer. Suppose there are positive numbers $\lambda_1,\ldots,\lambda_L$ such that $f(\w_1,\w_2,\ldots, \w_L)= f(\lambda_1\w_1,\lambda_2\w_2,\ldots, \lambda_L\w_L)$ for all $\w_l$. 
Then, with $\w= (\w_1,\w_2,\ldots, \w_L)$ and $\w^\lambda=(\lambda_1 \w_1,\lambda_2 \w_2,\ldots, \lambda_L \w_L)$, we have $\kappa^l(\w)=\kappa^l(\w^\lambda) \textrm{ for all } 1\leq l \leq L.$
\end{theorem}

We provide a proof in Appendix~\ref{sct:proofInvariance}.

\paragraph{An Averaging Alternative}

Experimental work \cite{hessianEigenvalueDensity} suggests that the spectrum of the Hessian has a lot of small values and only a few large outliers. We therefore consider the trace as an average of the spectrum to obtain $\kappa^l_\tau(\w):=||\vvec \w_l||^2\cdot Tr(H(\w_l))$ as a measure of flatness.
The same arguments as those used to prove Theorem~\ref{thm:invariance} also show the measure $\kappa^l_\tau$ to be independent with respect to the same layer-wise reparameterizations.

\section{Empirical Evaluation}
We empirically validate the practical usefulness of the proposed flatness measure by showing a strong correlation with the generalization error at local minima of the loss surface. 
For measuring the generalization error, we employ a Monte Carlo approximation of the target distribution defined by the testing dataset and measure the difference between loss value on this approximation and empirical error. In order to track the correlation of the flatness measure to the generalization error at local minima, sufficiently different minima should be achieved by training. The most popular technique is to train the model with small and large batch size \citep{scaleInvariantMeasure, keskarLarge, sensitivityGeneralization, identifyingGenProperties}, which we also employed.

\begin{figure}[h]
\minipage{0.5\textwidth}
  \includegraphics[width=\linewidth]{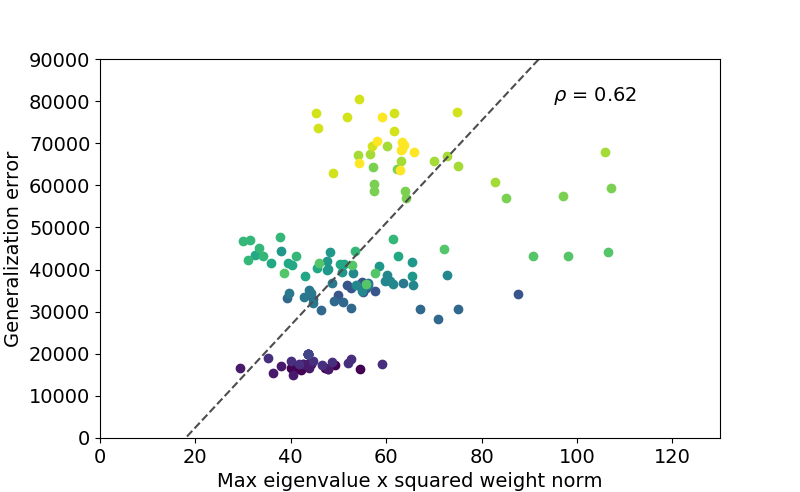}
\endminipage\hfill
\minipage{0.5\textwidth}
  \includegraphics[width=\linewidth]{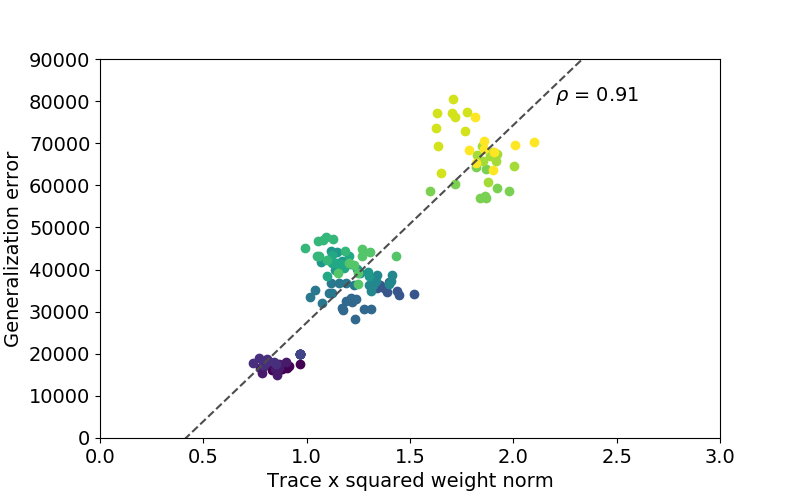}
\endminipage
\caption{LeNet5 characteristics after training on CIFAR10. Each color corresponds to a different setup of training, characterized by initialization strategy, mini batch size and learning rate. The setups are ordered in ascending order by the mini batch size, with the largest corresponding to the brightest color of the displayed points.}
\label{fig:flatness_cifar10}
\end{figure}

\begin{figure}[h]
\minipage{0.5\textwidth}
  \includegraphics[width=\linewidth]{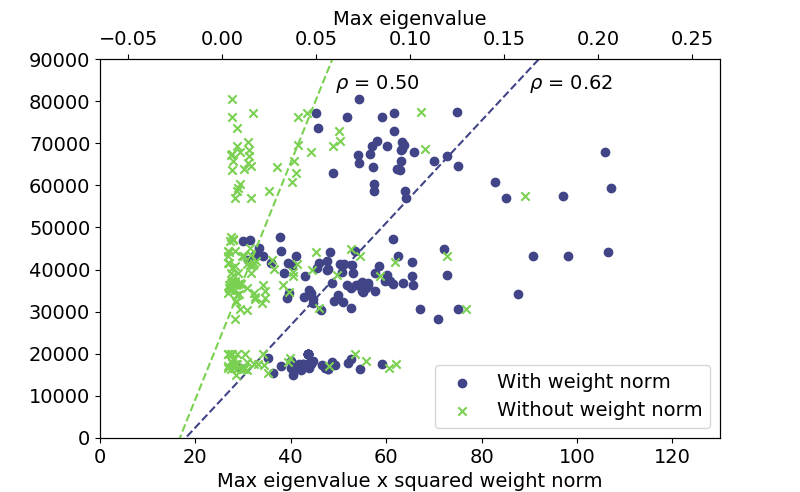}
\endminipage\hfill
\minipage{0.5\textwidth}
  \includegraphics[width=\linewidth]{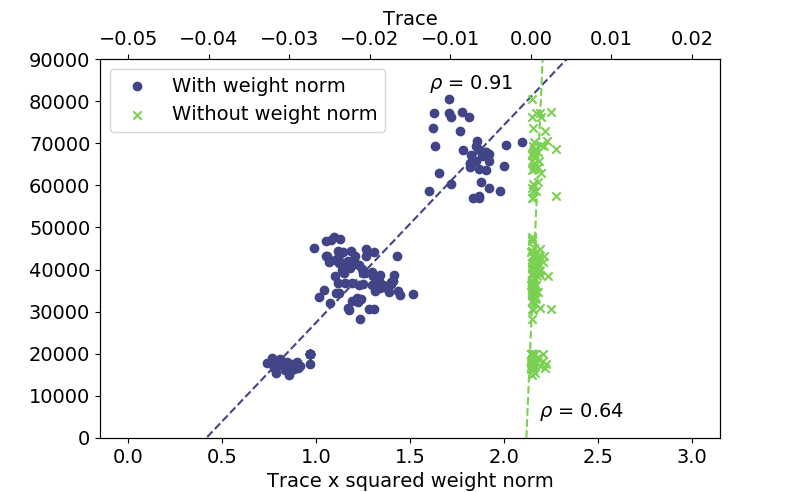}
\endminipage
\caption{LeNet5 configurations trained on CIFAR10 with random reparameterizations. Each figure displays both our measure and Hessian-based measure without weight norm multiplication. The correlation stays the same for the proposed measure, while Hessian-based measures fail to clearly indicate generalization error.}
\label{fig:reparam}
\end{figure}

A neural network (LeNet5~\citep{lecun2015lenet}) is trained on CIFAR10 multiple times until convergence with various training setups. This way, we obtain network configurations in multiple local minima. In particular four different initialization schemes were considered (Xavier normal, Kaiming uniform, uniform in $(-0.1,0.1)$, normal with $\mu = 0$ and $\sigma^2 = 0.1$), with four different mini-batch sizes ($4$, $32$, $64$, $512$) and corresponding learning rates to keep the ration between them equal ($0.001$, $0.008$, $0.02$, $0.1$) for the standard SGD optimizer. Each of the setups was run for $9$ times with different random initializations.  Here the generalization error is the difference between summed error values on test samples multiplied by $5$ (since the size of the training set is $5$ times larger) and summed error values on the training examples. Figure~\ref{fig:flatness_cifar10} shows the approximated generalization error with respect to the flatness measure (for both $\kappa^l$ and $\kappa^l_\tau$ with $l=5$ corresponding to the last hidden layer) for all network configurations. The correlation is significant for both measures, and it is stronger (with $\rho=0.91$) for $\kappa^5_\tau$. This indicates that taking into account the full spectrum of the Hessian is beneficial.
To investigate the invariance of the proposed measure to reparameterization, we apply the reparameterization discussed in Sec.~\ref{sec:measuresOfFlatness} to all networks using random factors in the interval $[5,25]$. The impact of the reparameterization on the proposed flatness measures in comparison to the traditional ones is shown in Figure~\ref{fig:reparam}. While the proposed flatness measures are not affected, the ones purely based on the Hessian have very weak correlation with the generalization error after the modifications.
Additional experiments conducted on MNIST dataset are described in Appendix~\ref{appndix:add_exp}.

In contrast to existing measures of flatness, our proposed measure is invariant to layer-wise reparameterizations of ReLU networks. However, we note that other reparameterizations are possible, e.g., we can use the positive homogeneity and multiply all incoming weights into a single neuron by a positive number $\lambda>0$ and multiply all outgoing weights of the same neuron by $\sfrac 1\lambda$. Our proposed measures of flatness $\kappa^l$ and $\kappa^l_\tau$ are in general not invariant to such reparameterizations. We define other flatness measures that are invariant to such reparameterizations as well in Appendix~\ref{appendix:additionalFlatnessMeasures}. 

Taking things together, we proposed a novel and practically useful flatness measure that strongly correlates with the generalization error while being invariant to reparameterization.


\newpage
\bibliography{featureRobustness}

\newpage
\appendix

\section{Proof of Theorem~\ref{thm:invariance}}\label{sct:proofInvariance}
In this section, we discuss the proof to Theorem~\ref{thm:invariance}. Before starting with the formal proof, we discuss the idea in a simplified setting to separate the essential insight from the complicated notation in the setting of neural networks.

Let $F,\tilde F:\RR^d\rightarrow \RR$ denote twice differentiable functions such that $F(w)=\tilde F(\lambda w)$ for all $w$ and all $\lambda>0$. Later, $w$ will correspond to weights of a specific layer of the neural network and the functions $F$ and $\tilde F$ will correspond respectively to the neural network functions before and after reparameterizations of possibly all layers of the network. We show that  
\[ \frac 1{\lambda^2}H(F(w))=H(\tilde F(\lambda w)).\]
Indeed, the second derivative of $\tilde F$ at $\lambda w$ with respect to coordinates $w_i,w_j$ is given by the differential quotient as
\begin{equation*}
\begin{split}
 \frac{\partial^2 \tilde F(\lambda w)}{\partial w_i \partial w_j}&=\lim_{h\rightarrow 0} \frac{\tilde F(\lambda w+he_i+he_j)-\tilde F(\lambda w+he_i)-\tilde F(\lambda w + he_j)+\tilde F(\lambda w)}{h^2} \\
&= \lim_{h\rightarrow 0} \frac{\tilde F(\lambda (w+\frac h\lambda e_i +\frac h\lambda e_j))-\tilde F(\lambda (w+\frac h\lambda e_i))-\tilde F(\lambda (w+\frac h\lambda e_j))+\tilde F(\lambda w)}{\left (\frac{h}{\lambda}\right ) ^2\lambda^2} \\
&= \frac{1}{\lambda^2}\lim_{h\rightarrow 0} \frac{F(w+\frac h\lambda e_i+ \frac h\lambda e_j)-F( w+\frac h\lambda e_i)-F( w+\frac h\lambda e_j)+ F(w)}{\left (\frac{h}{\lambda}\right ) ^2} \\
 & = \frac{1}{\lambda^2} \frac{\partial^2 F(w)}{\partial w_i \partial w_j}. \\
\end{split}
\end{equation*}

Since this holds for all combinations of coordinates, we see that $H\tilde F(\lambda w)= \sfrac{1}{\lambda^2}HF(w)$ for the Hessians of $F$ and $\tilde F$, and hence
\[||\lambda w||^2 H\tilde F(\lambda w)=\lambda^2 ||w||^2 \frac{1}{\lambda^2}HF(w)=||w||^2 HF(w).\]

\paragraph{Formal Proof of Theorem~\ref{thm:invariance}}

We are given a neural network function $f(x; \w_1,\w_2,\ldots, \w_L)$ parameterized by weights $\w_i$ of the $i$-th layer and positive numbers $\lambda_1,\ldots,\lambda_L$ such that $f(x; \w_1,\w_2,\ldots, \w_L)= f(x; \lambda_1\w_1,\lambda_2\w_2,\ldots, \lambda_L\w_L)$ for all $\w_i$ and all $x$. With $\w$ defined by $\w= (\w_1,\w_2,\ldots, \w_L)$,  $\w^\lambda_l=\lambda_l\w_l$ and $\w^\lambda=(\w^\lambda_1,\w^\lambda_2,\ldots, \w^\lambda_L)$, we aim to show that
\[\kappa^l(\w)=\kappa^l(\w^\lambda) ,\]
where $\kappa^l(\w)= || \w_l||^2\ \lambda_{max}^{H,l}(\w_l )$ is the product of the squared norm of vectorized weight matrix $\w_l$ with the maximal eigenvalue of the Hessian of the empirical error at $\w$ with respect to parameters $\w_l$.

Let $F(\uu):=\sum_{(x,y)\in S} \loss(f(x; \w_1,\w_2,\ldots,\uu,\ldots,\w_L),y)$ denote the loss as a function on the parameters of the $l$-th layer before reparameterization. Further, we let $\tilde F(\vv):=\sum_{(x,y)\in S} \loss( f(x;  \w^\lambda_1, \w_2^\lambda,\ldots,\vv,\ldots,\w^\lambda_L),y)$  denote the loss as a function on the parameters of the $l$-th layer after reparameterization. We define a linear function $\eta$ by $\eta(\uu)=\lambda_l \uu$. By assumption, we have that $\tilde F(\eta(\w_l))=F(\w_l)$ for all $\w_l$. By the chain rule, we compute for any variable $u^{(i,j)}$ of $\uu$,
\[ \frac{\partial F(\uu)}{\partial u^{(i,j)}}\Bigr|_{\uu=\w_l} 
= \frac{\partial \tilde F(\eta(\uu))}{\partial u^{(i,j)}}\Bigr|_{\uu=\w_l}\]
\[= \sum_{k,m} \frac{\partial \tilde F(\eta(\uu))}{\partial (\eta(\uu)^{(k,m)})}\Bigr|_{\eta(\uu)=\eta(\w_l)} \cdot \frac{\partial (\eta(\uu)^{(k,m)})}{\partial u^{(i,j)}}\Bigr|_{\eta(\uu)=\eta(\w_l)} \]
\[=  \frac{\partial \tilde F(\vv)}{\partial v^{(i,j)}}\Bigr|_{\vv=\lambda_l \w_l}\cdot \lambda_l. \]
Similarily, for second derivatives, we get for all $i,j,s,t$
\[ \frac{\partial^2 F(\uu)}{\partial u^{(i,j)}\partial u^{(s,t)} }\Bigr|_{\uu=\w_l} = \lambda_l^2  \frac{\partial \tilde F(\vv)}{\partial v^{(i,j)}\partial v^{(s,t)}}\Bigr|_{\vv=\lambda_l \w_l} \]

Consequently, the Hessian $H$ of the empirical error before reparameterization and the Hessian $\tilde H$ after reparameterization satisfy 
$H(\w_l,S) =\lambda_l ^2 \cdot \tilde H(\lambda_l \w_l ,S)$ and also $\lambda_{max}^{H,l}(\w_l)=\lambda_l^2 \cdot \lambda_{max}^{\tilde H,l}(\lambda_l \w_l )$. Therefore,
\[ \kappa^l(\w)=||\w_l||^2 \lambda_{max}^{H,l}(\w_l )=||\w_l||^2 \lambda_l^2\cdot  \lambda_{max}^{\tilde H,l}(\lambda_l \w_l)=||\lambda_l \w||^2  \lambda_{max}^{\tilde H,l}(\lambda_l \w_l)=\kappa^l(\w^\lambda).\]

\section{Additional Measures of Flatness}
\label{appendix:additionalFlatnessMeasures}

We present additional measures of flatness we have considered during our study. The original motivation to study additional measures was given by the observation that there are other possible reparameterizations of a fully connected ReLU network than suitable multiplication of layers by positive scalars: We can use the positive homogeneity and multiply all incoming weights into a single neuron by a positive number $\lambda>0$ and multiply all outgoing weights of the same neuron by $\sfrac 1\lambda$. Our previous measures of flatness $\kappa^l$ and $\kappa^l_\tau$ are in general not independent of the latter reparameterizations. We define for each layer $l$ and neuron $j$ in that layer a flatness measure by
\[\rho^l(j)(\w_{\ast}):= \w_{l \ast}(j)^T H\Ecal_{emp}(\w_{l \ast}(j)) \w_{l \ast}(j)\]
For each $l$ and $j$, this measure is invariant under all linear reparameterizations that do not change the network function. The proof of the following theorem is given in Section~\ref{sct:proofInvarianceAll}

\begin{theorem}\label{thm:invarianceAll}
Let $f=f(\w_1,\w_2,\ldots, \w_L)$ denote a neural network function parameterized by weights $\w_i$ of the $i$-th layer. Suppose there are positive numbers $\lambda^{(i,j)}_1,\ldots,\lambda^{(i,j)}_L$ such that the products $\w^\lambda_l$ obtained from multiplying weight $w^{(i,j)}_l$ at matrix position $(i,j)$ in layer $l$ by $\lambda^{(i,j)}_l$ satisfy that $f(\w_1,\w_2,\ldots, \w_L)= f(\w^\lambda_1,\w^\lambda_2,\ldots, \w^\lambda_L)$ for all $\w_i$. Then $\rho^l(j)(\w)=\rho^l(j)(\w^\lambda)$ for each $j$ and $l$.
\end{theorem}

We define a measure of flatness for a full layer by combinations of the measures of flatness for each individual neuron
\[\rho^l(\w_{\ast}):=\max_j \rho^l(j)(\w_{\ast}) \textrm{ and } \rho^l_{\sigma}(\w_{\ast}):=\sum_j \rho^l(j)(\w_{\ast})\]
Since each of the individual expressions is invariant under all linear reparameterizations, so are the maximum and sum.

\bgroup
\def\arraystretch{1.3}
\begin{table}
  \caption{Hessian based measures of flatness}
  \label{tbl:flatnessMeasures}
  \centering
  \begin{tabular}{lllll}
    \toprule
    Notation     & Definition     & One value per & Invariance  \\
    \midrule
$\kappa^l$ & $||\vvec \w_{l}||^2\cdot \lambda_{max}^{H,l}(\vvec{\w}_{l})$ & layer & layer-wise mult. by pos scalar\\
$\kappa^l_\tau$ & $||\vvec{\w}_{l}||^2\cdot  Tr(H\Ecal_{emp}(\vvec{\w}_{l},S))$ & layer & layer-wise mult. by pos scalar\\
$\kappa^{max}$ & $\max_l \kappa^l(\w)$ & network & layer-wise mult. by pos scalar\\
$\kappa^{\Sigma}$ & $\sum_{l=1}^L \kappa^l(\w)$ & network &layer-wise mult. by pos scalar\\
$\kappa^{max}_{\tau}$ & $\max_l \kappa^l_\tau(\w)$ &network & layer-wise mult. by pos scalar\\
$\kappa^{\Sigma}_\tau$ &$\sum_{l=1}^L \kappa^l_\tau(\w)$ & network & layer-wise mult. by pos scalar\\
$\rho^l(j)$ & $ \w_{l}(j)^T H\Ecal_{emp}(\w_{l}(j),S) \w_{l}(j)$ & neuron & all linear reparameterizations\\
$\rho^l$ & $\max_j \rho^l(j)(\w)$ & layer & all linear reparameterizations\\
$\rho^l_{\sigma}$ & $\sum_j \rho^l(j)(\w)$ & layer & all linear reparameterizations\\
$\rho^{max}$ & $\max_l \rho^l(\w)$ & network & all linear reparameterizations \\
$\rho^{\Sigma}$ & $\sum_{l=1}^L \rho^l_\sigma(\w)$ &network & all linear reparameterizations\\
   \bottomrule
  \end{tabular}
\end{table}

\paragraph{One Value for all Layers}

It is clear that a low value of $\kappa^l$ for a specific layer $l$ alone cannot explain good performance. We therefore consider simple common bounds by combinations of the individual terms $\kappa^l$, e.g., by taking the maximum of $\kappa_l$ over all layers, $\kappa^{max}(\w_\ast):=\max_l \kappa^l(\w_\ast)$, or the sum $\kappa^{\Sigma}(\w_\ast):=\sum_{l=1}^L \kappa^l(\w_\ast)$. Since each of the individual expressions are invariant under linear reparameterizations of full layers, so are the maximum and sum.

Finally, we define $\rho^{max}(\w_\ast):=\max_l \rho^l(\w_\ast)$ and $\rho^{\Sigma}(\w_\ast):=\sum_{l=1}^L \rho^l_\sigma(\w_\ast)$.

Table~\ref{tbl:flatnessMeasures} summarizes all our measures of flatness, specifying whether each measure is defined per network, layer or neuron, and whether it is invariant layer-wise multiplication by a positive scalar (as considered in Theorem~\ref{thm:invariance}) or invariant under all linear reparameterization (as considered in Theorem~\ref{thm:invarianceAll}).

\subsection{Proof of Theorem~\ref{thm:invarianceAll}}\label{sct:proofInvarianceAll}

As in Subsection~\ref{sct:proofInvariance}, we first present the idea in a simplified setting. 

For the proof of Theorem~\ref{thm:invarianceAll} we need to consider the case when we multiply coordinates by different scalars. Let $F:\RR^{2}\rightarrow \RR$ denote twice differentiable functions such that $F(v,w)= \tilde F(\lambda v,  \mu w)$ for all $v\in \RR$, $w\in \RR$ and all $\lambda,\mu>0$. In the formal proof, $v,w$ will correspond to two outgoing weights for a specific neuron, while again $F$ and $\tilde F$ correspond to network functions before and after reparameterizations of all possibly all weights of the neural network. Then 
\[ (v,w) \cdot HF(v,w)\cdot \left ( \begin{array}{c} v \\ w\end{array}\right )=(\lambda v, \mu w) \cdot HF(\lambda v,\mu w)\cdot \left ( \begin{array}{c} \lambda v \\ \mu w\end{array}\right )\]
for all $v,w$ and all $\lambda,\mu>0$.

Indeed, the second derivative of $\tilde F$ at $(\lambda v, \mu w)$ with respect to coordinates $v,w$ is given by the differential quotient as
\begin{equation*}
\begin{split}
 \frac{\partial^2  \tilde F(\lambda v, \mu w )}{\partial v \partial w}  &=\lim_{h,k\rightarrow 0} \frac{ \tilde F(\lambda v +h,\mu w+k e)- \tilde F(\lambda v+h,\mu w)- \tilde F(\lambda v,\mu w+k )+ \tilde F(\lambda v,w)}{hk}\\
&\hspace{-1cm}= \lim_{h,k\rightarrow 0} \frac{\tilde F(\lambda (v+\frac h\lambda ),\mu(w+\frac k\mu ))- \tilde F(\lambda (v+\frac h\lambda ,\mu w))- \tilde F(\lambda v,\mu (w+\frac k\lambda ))+ \tilde  F(\lambda v, \mu w)}{\left (\frac{h}{\lambda}\right )\left (\frac k\mu\right ) \lambda\mu } \\
&\hspace{-1cm}= \frac{1}{\lambda\mu}\lim_{h,k\rightarrow 0} \frac{F(v+\frac h\lambda ,w+\frac k\mu )- F( v+\frac h\lambda ,w)- F( v,w+\frac k\mu )+ F(v,w)}{\frac{h}{\lambda}\frac{k}{\mu}} \\
&\hspace{-1cm}= \frac{1}{\lambda\mu} \frac{\partial^2 F(v,w)}{\partial v \partial w}. \\
\end{split}
\end{equation*}

From the calculation above, we also see that 
\[ \frac{\partial^2  \tilde F(\lambda v, \mu w )}{\partial v \partial v}= \frac{1}{\lambda^2} \frac{\partial^2 F(v,w)}{\partial v \partial v}\textrm{, and }
 \frac{\partial^2  \tilde F(\lambda v, \mu w )}{\partial w \partial w}= \frac{1}{\mu^2} \frac{\partial^2 F(v,w)}{\partial w \partial w}.\]
It follows that 
\begin{equation*}
\begin{split}
 (v,w)\cdot HF(v,w) \cdot \left ( \begin{array}{c} v \\ w\end{array}\right ) & =v^2  \frac{\partial^2 F(v,w)}{\partial v \partial v} + 2 vw \frac{\partial^2 F(v,w)}{\partial v \partial w}+ w^2 \frac{\partial^2 F(v,w)}{\partial w \partial w} \\
&=(\lambda v)^2  \frac{\partial^2 \tilde F(v,w)}{\partial v \partial v} + 2 (\lambda v)(\mu w) \frac{\partial^2 \tilde F(v,w)}{\partial v \partial w}+ (\mu w)^2 \frac{\partial^2 \tilde F(v,w)}{\partial w \partial w} \\
& = (\lambda v, \mu w) \cdot  HF(\lambda v,\mu w)\cdot \left ( \begin{array}{c} \lambda v \\ \mu w\end{array}\right ).
\end{split}
\end{equation*}

\paragraph{Formal Proof of Theorem~\ref{thm:invarianceAll}}

We are given a neural network function $f(x; \w_1,\w_2,\ldots, \w_L)$ parameterized by weights $\w_i$ of the $i$-th layer and positive numbers $\lambda^{(i,j)}_1,\ldots,\lambda^{(i,j)}_L$ such that the products $\w^\lambda_l$ obtained from multiplying weight $w^{(i,j)}_l$ at matrix position $(i,j)$ in layer $l$ by $\lambda^{(i,j)}_l$ satisfies that $f(x; \w_1,\w_2,\ldots, \w_L)= f(x; \w^\lambda_1,\w^\lambda_2,\ldots, \w^\lambda_L)$ for all $\w_i$ and all $x$. We aim to show that \[\rho^l(j)(\w)=\rho^l(j)(\w^\lambda)\] for each $j$ and $l$
where $\rho^l(j)(\w)=\w_{l}(j)^T H\Ecal_{emp}(\w_{l}(j),S) \w_{l}(j)$, $\w_l(j)$ denotes the $j$-th column of the weight matrix at the $l$-th layer and $H\Ecal_{emp}(\w_{l}(j),S)$ denotes the Hessian of the empirical error with respect to the weight parameters in $\w_l(j)$. Similar to the above, we denote by $\w_l(j)^\lambda$ the product obtained from multiplying weight $\w_l(j)_i=w^{(i,j)}_l$ at matrix position $(i,j)$ in layer $l$ by $\lambda^{(i,j)}$.

The proof is very similar to the proof of Theorem~\ref{thm:invariance}, only this time we have to take the different parameters $\lambda_l^{(i,j)}$ into account. For fixed layer $l$, we denote the $j$-th column of $\w_l$ and $\w_l(j)$.  

Let 
\begin{equation*}
    \begin{split}
    F(\uu):=\sum_{(x,y)\in S} \loss(f(x; \w_1,\w_2,&\ldots,[\w_l(1),\ldots,\w_l(j-1),\uu,\w_l(j+1),\ldots \w_l(n_l)],\\
    &\ldots,\w_L),y)
    \end{split}
\end{equation*}
denote the loss as a function on the parameters of the $j$-th column in the $l$-th layer before reparameterization and 
\begin{equation*}
    \begin{split}
        \tilde F(\vv):=\sum_{(x,y)\in S} \loss( f(x_i;  \w_1^{\lambda_1},\w_2^{\lambda_2} ,&\ldots,[\w_l(1)^\lambda ,\ldots,\w_l(j-1)^\lambda,\vv,\w_l(j+1)^\lambda,\ldots w_l(n_l)^\lambda], \\
        &\ldots, \w_L{^{\lambda_L}}),y)
    \end{split}
\end{equation*}
denote the loss as a function on the parameters of the $j$-th neuron in the $l$-th layer after reparameterization. 

We define a linear function $\eta$ by \[\eta(\uu)=\eta(u_1,u_2,\ldots u_{n_l})=\eta(u_1\lambda_l^{(1,j)},u_2\lambda_l^{(2,j)},\ldots u_{n_l}\lambda_l^{(n,j)}).\] By assumption, we have that $\tilde F(\eta(\w_l(j)))=F(\w_l(j))$ for all $\w_l(j)$. By the chain rule, we compute for any variable $u_i$ of $\uu$,
\[ \frac{\partial F(\uu)}{\partial u_i}\Bigr|_{\uu=\w_l(j)} 
= \frac{\partial \tilde F(\eta(\uu))}{\partial u_i}\Bigr|_{\uu=\w_l(j)}\]
\[= \sum_k \frac{\partial \tilde F(\eta(\uu))}{\partial (\eta(\uu)_k)}\Bigr|_{\eta(\uu)=\eta(\w_l(j))} \cdot \frac{\partial (\eta(\uu)_k)}{\partial u_i}\Bigr|_{\eta(\uu)=\eta(\w_l(j))} \]
\[=  \frac{\partial \tilde F(\vv)}{\partial v_i}\Bigr|_{\vv= \w_l(j)^\lambda}\cdot \lambda_l^{(i,j)}. \]
Similarily, for second derivatives, we get for all $i,s$,
\[ \frac{\partial^2 F(\uu)}{\partial u_i \partial u_s }\Bigr|_{\uu=\w_l(j)} = \lambda_l^{(i,j)} \lambda_l^{(s,j)}  \frac{\partial \tilde F(\vv)}{\partial v_i \partial v_j }\Bigr|_{\vv= \w_l(j)^\lambda}. \]

Consequently, the Hessian $HF$ of the empirical error before reparameterization and the Hessian $H\tilde F$ after reparameterization satisfy that at position $(i,s)$ of the Hessian matrix,
\[HF(\w_l)_{(i,s)} =\lambda_l ^{(i,j)} \lambda_l ^{(s,j)} \cdot H\tilde F( \w^\lambda_l)_{(i,s)}.\]  Therefore,
\begin{equation*}
    \begin{split}
 \rho^l(j)(\w)&=\w_l(j)^T \cdot HF(\w_l) \cdot \w_l(j) = \sum_{i,s} w_l^{(i,j)} w_l^{(s,j)}  HF(\w_l)_{(i,s)}\\
 &= \sum_{i,s} w_l^{(i,j)} w_l^{(s,j)} \lambda_l ^{(i,j)} \lambda_l ^{(s,j)} \cdot H\tilde F (\w_l^\lambda)_{(i,s)}\\
& = \sum_{i,s} \lambda_l ^{((i,j)} w_l^{i,j)} \lambda_l ^{(s,j)}  w_l^{(s,j)} \cdot H\tilde F(\w_l^\lambda)_{(i,s)}\\
&= (\w_l(j)^{\lambda})^T \cdot H\tilde F(\w_l^\lambda) \cdot \w_l(j)^\lambda = \rho^l(j)(\w^\lambda)  
    \end{split}
\end{equation*}

\section{Additional experiments}

\label{appndix:add_exp}

In addition to the evaluation on the CIFAR10 dataset with LeNet5 network, we also conducted experiments on the MNIST dataset. For learning with this data, we employed a custom fully connected network with ReLU activations containing $4$ hidden layers with $50$, $50$, $50$, and $30$ neurons correspondingly. The output layer has $10$ neurons with softmax activation. The networks were trained till convergence on the training dataset of MNIST, moreover, the configurations that achieved larger than $0.07$ training error were filtered out. All the networks were initialized according to Xavier normal scheme with random seed. For obtaining different convergence minima the batch size was varied between $1000$, $2000$, $4000$, $8000$ with learning rate changed from $0.02$ to $1.6$ correspondingly to keep the ratio constant. All the configurations were trained with SGD. Figure~\ref{fig:flatness_mnist} shows the correlation between the layer-wise flatness measure based on the trace of the Hessian for the corresponding layer. The values for all four hidden layers are calculated (the trace is not normalized) and aligned with values of generalization error (difference between normalized test error and train error). The observed correlation is strong (with $\rho \geq 0.7$) and varies slightly for different layers, nevertheless it is hard to identify the most influential layer for identifying generalization properties.

\begin{figure}[h]
   \centering
   \subfloat{\includegraphics[width=.45\textwidth]{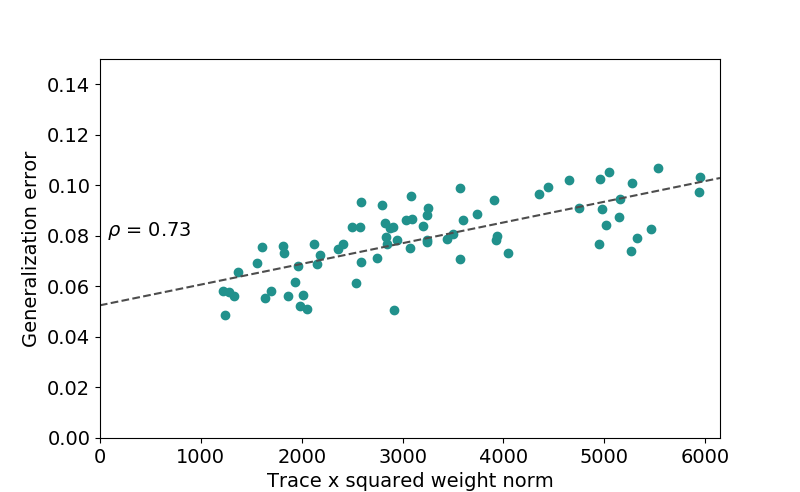}}\quad
   \subfloat{\includegraphics[width=.45\textwidth]{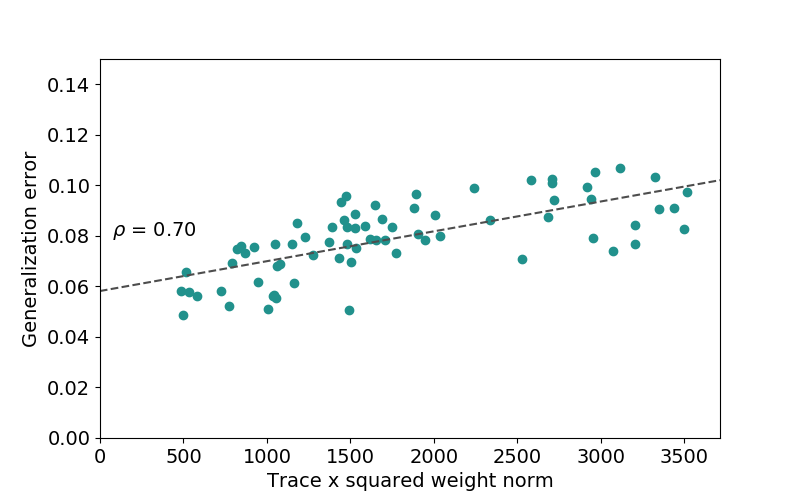}}\\
   \subfloat{\includegraphics[width=.45\textwidth]{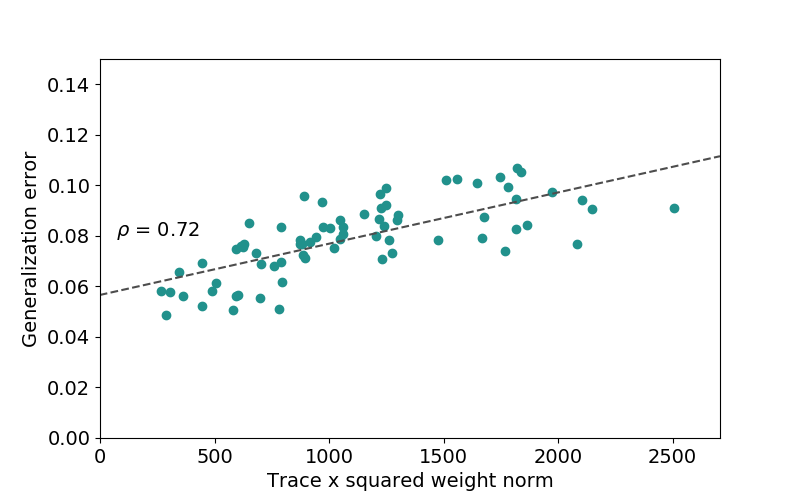}}\quad
   \subfloat{\includegraphics[width=.45\textwidth]{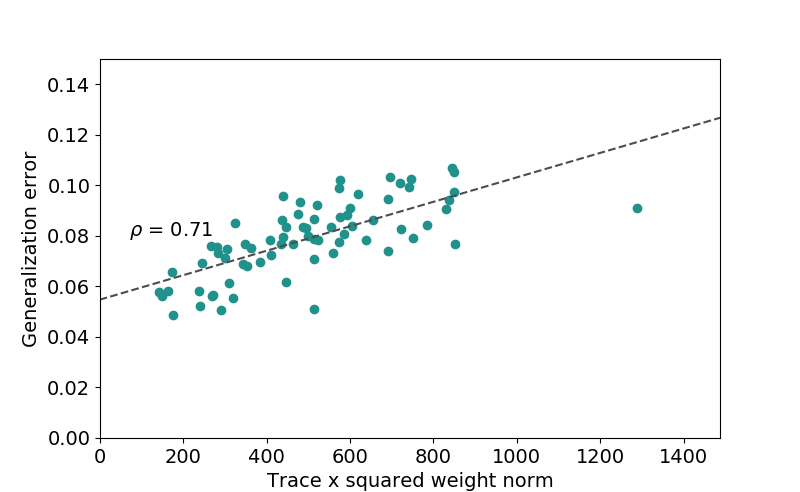}}
\caption{Layer-wise flatness measure calculated for MNIST trained fully-connected network. Four plots correspond to four hidden layers of the network. For each of the layers a strong correlation with generalization error can be observed.}
\label{fig:flatness_mnist}
\end{figure}

We also calculated neuron-wise flatness measures described in Appendix~\ref{appendix:additionalFlatnessMeasures} for these network configurations. In Figure~\ref{fig:flatness_neuronwise_mnist} we depicted correlation between $\rho_{\sigma}^l$ and generalization error for each of the layers, and in Figure~\ref{fig:flatness_neuronwise_max_mnist}--between $\rho^l$ and generalization error. The observed correlation is again significant, but compared to the previous measure we can see that it might differ considerably depending on the layer.  

\begin{figure}[h]
   \centering
   \subfloat{\includegraphics[width=.45\textwidth]{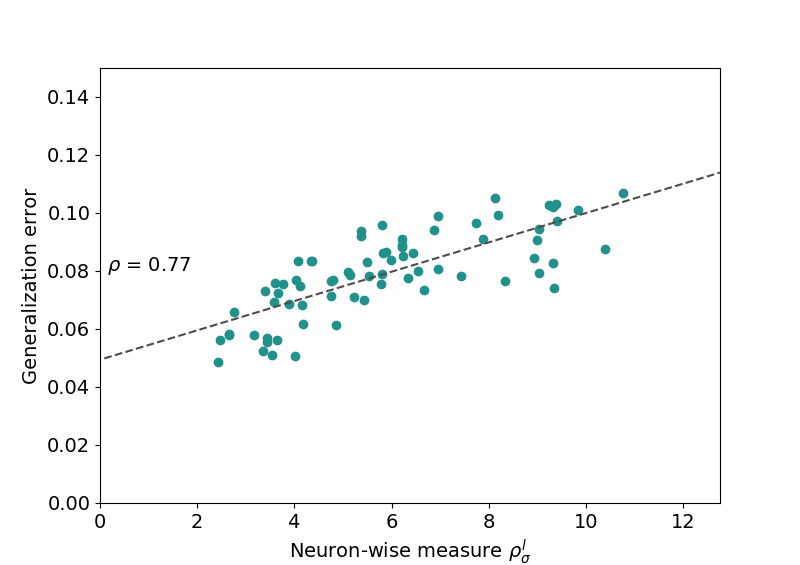}}\quad
   \subfloat{\includegraphics[width=.45\textwidth]{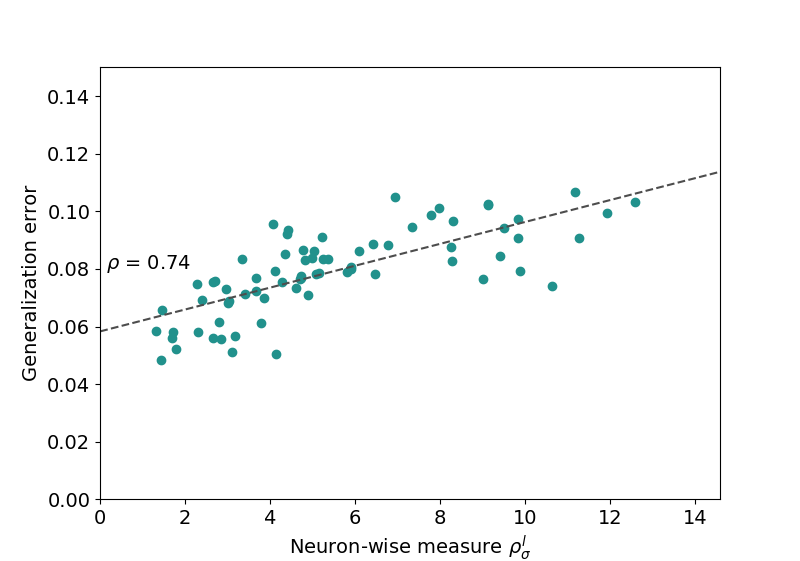}}\\
   \subfloat{\includegraphics[width=.45\textwidth]{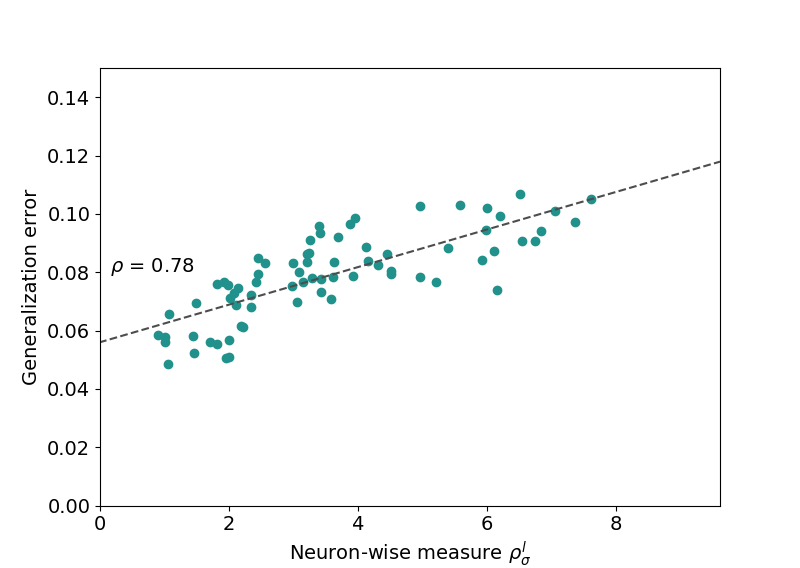}}\quad
   \subfloat{\includegraphics[width=.45\textwidth]{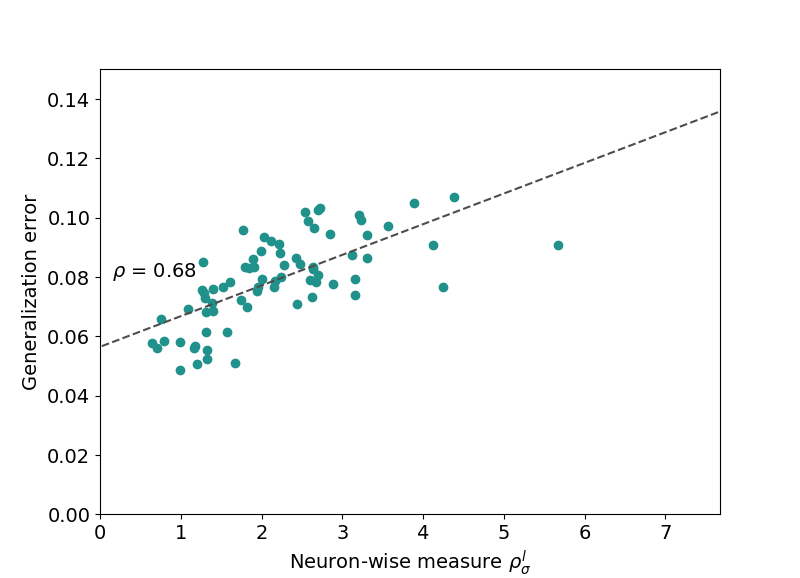}}
\caption{Neuron-wise flatness measure $\rho_{\sigma}^l$ calculated for each of the hidden layers for the fully-connected network trained on MNIST dataset. Each plot corresponds to a layer.}
\label{fig:flatness_neuronwise_mnist}
\end{figure}

\begin{figure}[h]
   \centering
   \subfloat{\includegraphics[width=.45\textwidth]{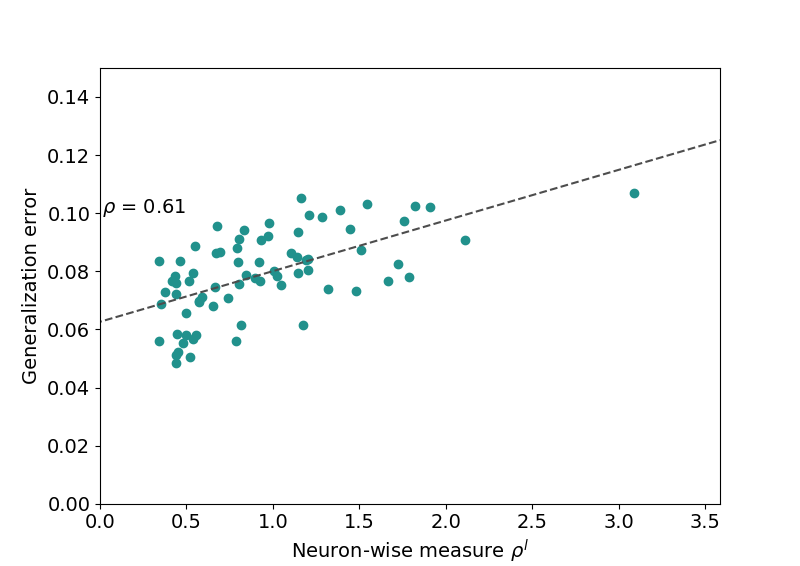}}\quad
   \subfloat{\includegraphics[width=.45\textwidth]{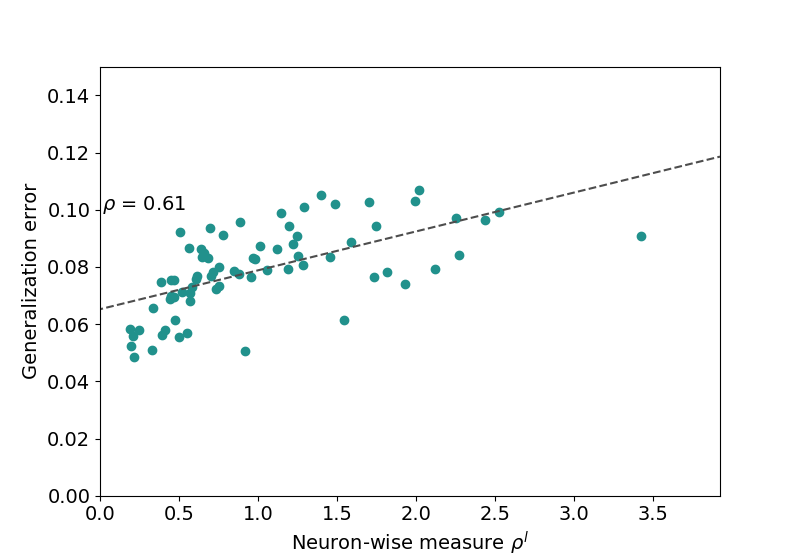}}\\
   \subfloat{\includegraphics[width=.45\textwidth]{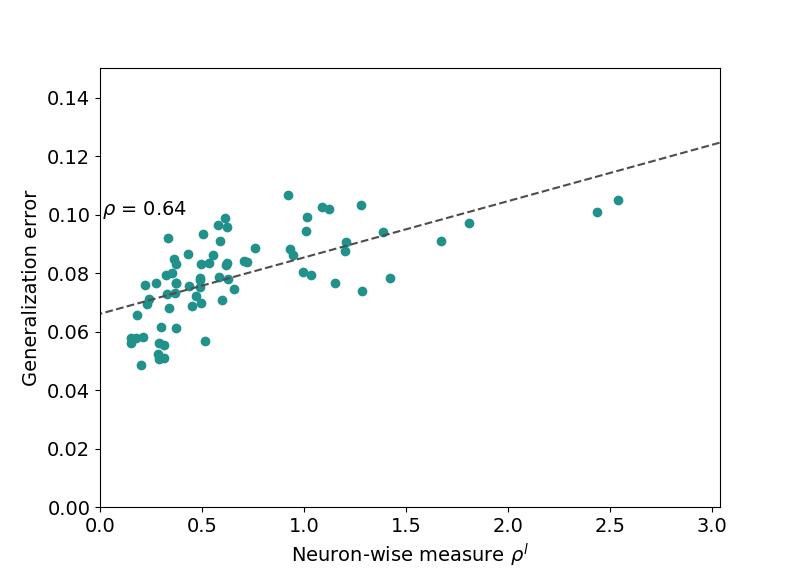}}\quad
   \subfloat{\includegraphics[width=.45\textwidth]{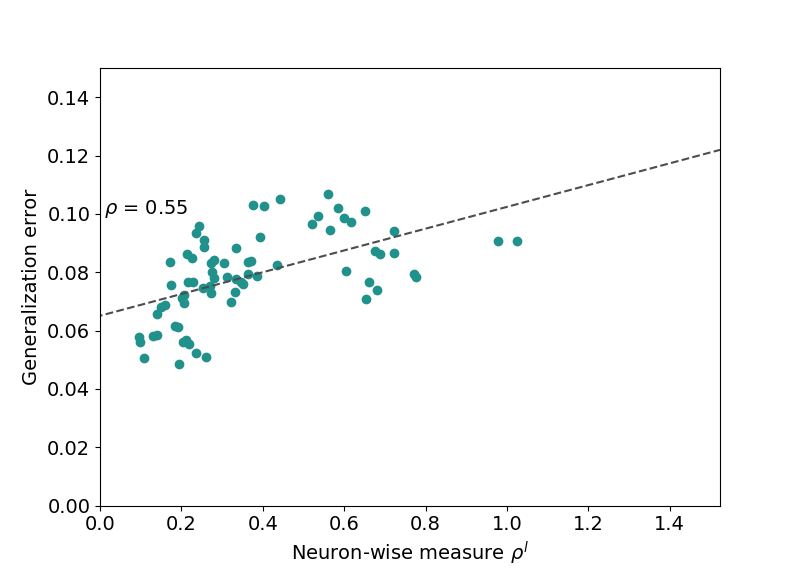}}
\caption{Neuron-wise flatness measure $\rho^l$ calculated for each of the hidden layers for the fully-connected network trained on MNIST dataset. Each plot corresponds to a layer.}
\label{fig:flatness_neuronwise_max_mnist}
\end{figure}

The network-wise flatness measures can be based both on layer-wise and neuron-wise measures as defined in Appendix~\ref{appendix:additionalFlatnessMeasures}. We computed $\kappa_{\tau}^{max}$, $\kappa_{\tau}^{\Sigma}$, $\rho^{max}$, and $\rho^{\Sigma}$ and depicted them in Figure~\ref{fig:flatness_networkwise_mnist}. Interesting to note, that each of the network-wise measures has a larger correlation with generalization loss than the original neuron-wise and layer-wise measures.

\begin{figure}[h]
   \centering
   \subfloat{\includegraphics[width=.45\textwidth]{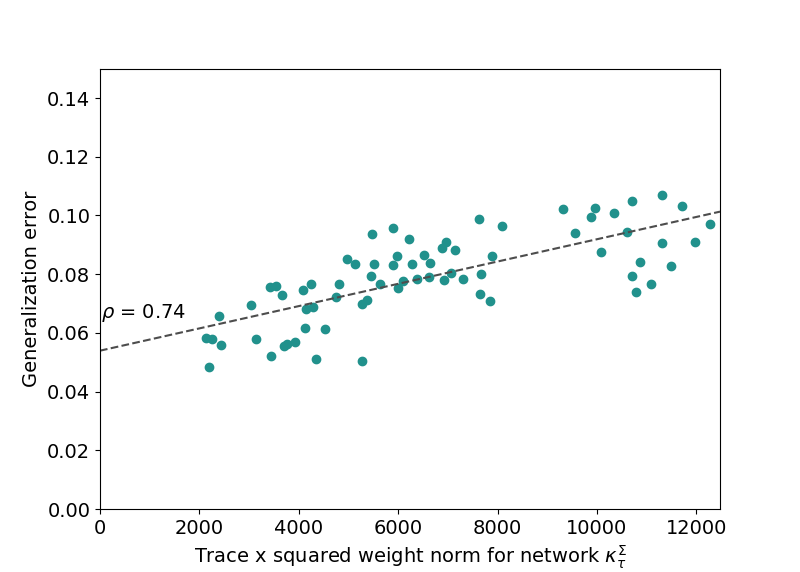}}\quad
   \subfloat{\includegraphics[width=.45\textwidth]{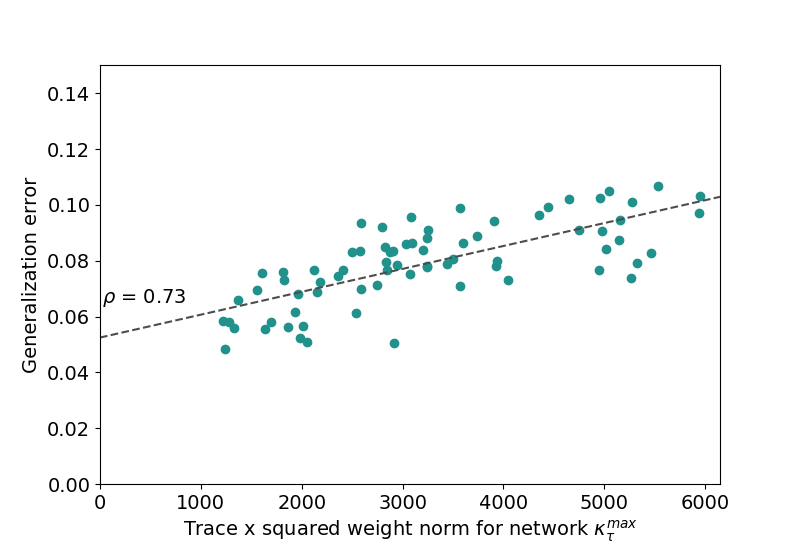}}\\
   \subfloat{\includegraphics[width=.45\textwidth]{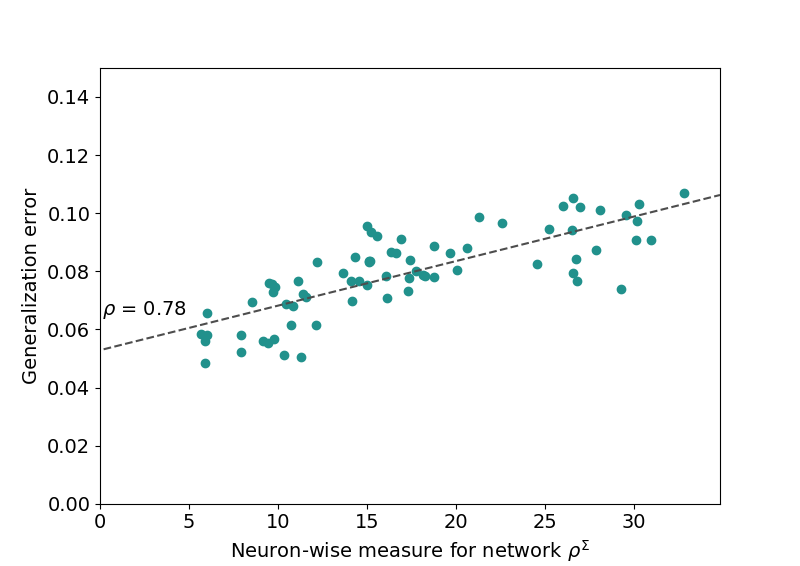}}\quad
   \subfloat{\includegraphics[width=.45\textwidth]{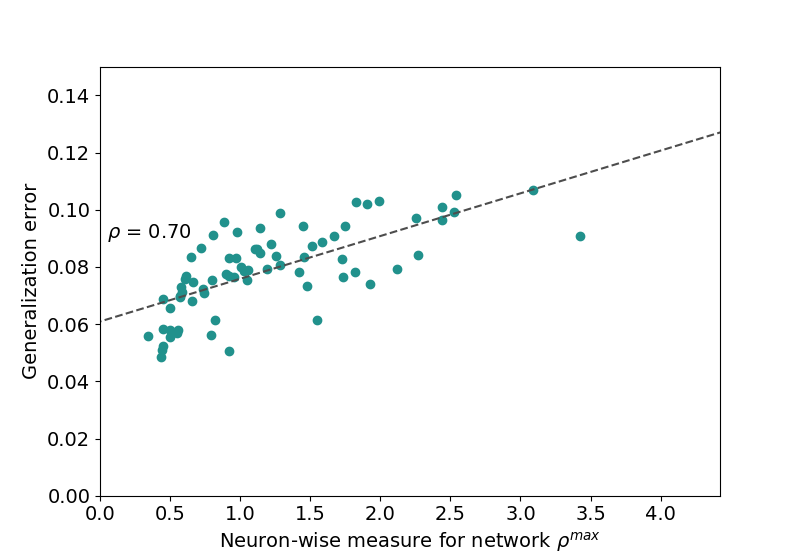}}
\caption{Network-wise flatness measures based on various neuron-wise and trace layer-wise measures for the fully-connected network trained on MNIST dataset.}
\label{fig:flatness_networkwise_mnist}
\end{figure}

\subsection{Proof of Equation~\eqref{eq:bound}}
First note that for $||A||\leq 1$,
\begin{equation}\label{eq:Frobenius}
\begin{split}
||wA||_F&=\left |\left |\left ( w_1|w_2|\ldots|w_m\right)A \right |\right|_F = \left |\left |\left ( w_1A|w_2A|\ldots|w_mA\right ) \right |\right |_F \\
&= \sqrt{ \sum_{j=1}^m ||w_jA||_2^2}\leq \sqrt{ \sum_{j=1}^m ||w_j||_2^2}\\
&=||w||_F .\\
\end{split}
\end{equation}

From \eqref{eq:calculation1} and  \eqref{eq:calculation2} we get
\begin{equation*}
\begin{split}
\max_{||A||\leq 1} \mathcal{F}(\delta,S,A) & \stackrel{\eqref{eq:calculation1}, \eqref{eq:calculation2}}{=} \max_{||A||\leq 1} \frac{\delta^2}{2} \langle   \vvec{\w_\ast A} ,\ H\Ecal_{emp}(\vvec \w_\ast,S) \cdot ( \vvec{\w_\ast A}) \rangle + \mathcal{O}(\delta^3)\\
& \stackrel{\eqref{eq:Frobenius}}{\leq} \max_{||\mathbf{z}||_2\leq ||w_\ast||_F }  \frac{\delta^2}{2}  \mathbf{z}^T  H\Ecal_{emp}(\vvec \w_\ast,S) \mathbf{z} + \mathcal{O}(\delta^3)\\
&=\max_{||\mathbf{z}||_2 = 1} \frac{\delta^2}{2} ||w_\ast||_F^2\  \mathbf{z}^T  H\Ecal_{emp}(\vvec \w_\ast,S) \mathbf{z}+ \mathcal{O}(\delta^3) \\
&=\frac{\delta^2}{2} ||\vvec{\w}_\ast||_F^2\  \lambda^H_{max}(\vvec{\w}_\ast) + \mathcal{O}(\delta^3),\\
\end{split}    
\end{equation*}
where we used the identity that $\max_{||x||=1}x^TMx=\lambda^M_{max}$ for any symmetric matrix $M$.

\section{Additional properties of feature robustness}

\subsection{Relation to noise injection at the feature layer}\label{app:noise}
Feature robustness is related to noise injection in the layer of consideration. By defining a probability measure $\mathcal{P}_A$ on matrices $A\in\RR^{m\times m}$ of norm $||A||\leq 1$, we can take expectations over matrices. An expectation over such matrices induces for each sample $x\in\Xcal$ an expectation over a probability distribution of vectors $\xi\in \RR^m$ with $||\xi|| \leq ||\phi(x)||$. We find the induced probability distribution $\mathcal{P}_x$ from the measure $P_x$ defined by 
$P_x(T)=\mathcal{P}_A(\{A\ |\ A\phi(x)\in T\})$ for a measurable subset $T\subseteq \RR^m$. Then,
\begin{equation*}
\begin{split}
    \Exp{A\sim \mathcal{P}_A}{\mathcal{F} (\delta, S, A)}&= \Exp{A\sim \mathcal{P}_A}{ \frac{1}{|S|} \sum_{(x,y)\in S} \left [ \loss(\psi( \phi(x) + \delta A\phi(x),y ))-\loss(f(x),y) \right] } \\
 &= \frac{1}{|S|} \sum_{(x,y)\in S} \Exp{\xi_x\in \mathcal{P}_x}{\  \loss(\psi( \phi(x) + \delta \xi_x )-\loss(f(x),y)\ }.
\end{split}
\end{equation*}
The latter is robustness to noise injection according to noise distribution $\mathcal{P}_x$ for sample $x$ at the feature layer defined by $\phi$.

\subsection{Adversarial examples}\label{app:adversarial}

\paragraph{Large changes of loss (adversarial examples) can be hidden in the mean in the definition of feature robustness.} 

We have seen that flatness of the loss curve with respect to some weights is related to the mean change in loss value when perturbing all data points $x_i$ into directions $Ax_i$ for some matrix $A$. For a common bound over different directions governed by the matrix $A$, we restrict ourselves to matrices $||A||\leq 1$. One may therefore wonder, what freedom of perturbing individual points do we have?

At first, note that for each fixed sample $x_{i_0}$ and direction $z_{i_0}$ there is a matrix $A$ such that $Ax_{i_0}=z_{i_0}$, so each direction for each datapoint can be considered within a bound as above. We get little insight over the change of loss for this perturbation however, since a larger change of the loss may go missing in the mean change of loss over all data points considered in the same bound.

The bound involving $\kappa(\w_\ast)$ from above does not directly allow to check the change of the loss when perturbing the samples $x_i$ independently into arbitrary directions . For example, suppose we have two samples close to each other and we are interested in the change of loss when perturbing them into directions orthogonal to each other. Specifically, suppose our dataset contains the points $(1,0,0, \ldots,0)$ and $(1,\epsilon,0, \ldots,0)$ for some small $\epsilon$, and we aim to check how the loss changes when perturbing $(1,0,0, \ldots,0)$ into direction $(1,0,0,\ldots,0)$ and $(1,\epsilon,0, \ldots,0)$ orthogonally into direction $(0,1,0,\ldots,0)$. To allow for this simultaneous change, our matrix $A$ has to be of the form \[A=\left ( \begin{array}{ccc} 1& -\frac 1\epsilon & \ldots \\ 0&\frac 1\epsilon & \ldots \\ 0 & \\ \vdots & \vdots \\ 0 & 0 & \ldots \end{array}\right ).\]
Then \[||A||\geq ||A\cdot \left (\begin{array}{c} 0\\1\\0\\ \vdots,\\ 0\end{array} \right )|| =||(-\frac 1\epsilon, \frac 1\epsilon, 0,\ldots)|| =\frac{\sqrt{2}}{\epsilon}.\]
Hence, our desired alterations of the input necessarily lead to a large matrix norm $||A||$ and our attainable bound with $||A||^2 \kappa(\w_\ast)$ becomes almost vacuous.

\subsection{Convolutional Layers}\label{app:convolution}
Feature robustness is not restricted to fully connected neural networks. In this section, we briefly consider convolutional layers $\w \ast x$. Using linearity, we get  $\w\ast (x+\delta x)=(\w+\delta \w)\ast x.$
What about changes $(\w+\delta \w A)$ for some matrix $A$? Since convolution is a linear function, there is a matrix $W$ such that $\overrightarrow{\w\ast x}=W\vvec x$ and there is a matrix $W_A$ such that $\overrightarrow{ \w A\ast x}=W_A \vvec x$. We assume that the convolutional layer is dimensionality-reducing, $W\in \RR^{n\times m}, m<n$ and that the matrix $W$ has full rank, so that there is a matrix $V$ with $WV=I_m$.\footnote{This holds for example for a convolutional filter with stride one without padding, as in this case $W$ has a Toeplitz submatrix of size $(m\times m)$.}
Then 
\[(\w+\delta \w A)\ast x= W\vvec x+\delta W_A\vvec x = W\vvec x+\delta W V W_B \vvec x= W(\vvec x+\delta V W_B\vvec x).\]
As a consequence, similar considerations of flatness and feature robustness can be considered for convolutional layers.

\end{document}